\documentclass[journal,twocolumn]{IEEEtran}
\usepackage{epsfig}
\usepackage{graphicx}
\usepackage{amsmath}
\usepackage{amssymb,subfigure,cases}
\usepackage{setspace}

\usepackage{color}
\usepackage{url}
\usepackage{makecell,multirow,diagbox}

\usepackage{indentfirst}

\ifCLASSINFOpdf

\else

\fi

\definecolor{darkblue}{rgb}{0.0,0.0,1.0}

\begin{document}

\title{GETNET: A General End-to-end Two-dimensional CNN Framework for Hyperspectral Image Change Detection}

\author {  
		Qi~Wang,~\IEEEmembership{Senior Member,~IEEE},    
		Zhenghang~Yuan,
		Qian Du,~\IEEEmembership{Fellow,~IEEE},
		and Xuelong Li,~\IEEEmembership{Fellow,~IEEE}     
\IEEEcompsocitemizethanks{\IEEEcompsocthanksitem This work was supported by the National Key R\&D Program of China under Grant 2017YFB1002202, National Natural Science Foundation of China under Grant 61773316, Fundamental Research Funds for the Central Universities under Grant 3102017AX010, and the Open Research Fund of Key Laboratory of Spectral Imaging Technology, Chinese Academy of Sciences.\protect
	
2019 IEEE. Personal use of this material is permitted. Permission from IEEE must be obtained for all other uses, in any current or future media, including reprinting/republishing this material for advertising or promotional purposes, creating new collective works, for resale or redistribution to servers or lists, or reuse of any copyrighted component of this work in other works.
}
}

\markboth{IEEE TRANSACTIONS ON GEOSCIENCE AND REMOTE SENSING}%
{Shell \MakeLowercase{\textit{et al.}}: Bare Advanced Demo of IEEEtran.cls for Journals}


\maketitle

\begin{abstract}
Change detection (CD) is an important application of remote sensing, which provides timely change information about large-scale Earth surface. With the emergence of hyperspectral imagery, CD technology has been greatly promoted, as hyperspectral data with high spectral resolution are capable of detecting finer changes than using the traditional multispectral imagery. Nevertheless, the high dimension of hyperspectral data makes it difficult to implement traditional CD algorithms. Besides, endmember abundance information at subpixel level is often not fully utilized. In order to better handle high dimension problem and explore abundance information, this paper presents a General End-to-end Two-dimensional CNN (GETNET) framework for hyperspectral image change detection (HSI-CD). The main contributions of this work are threefold: 1) Mixed-affinity matrix that integrates subpixel representation is introduced to mine more cross-channel gradient features and fuse multi-source information; 2) 2-D CNN is designed to learn the discriminative features effectively from multi-source data at a higher level and enhance the generalization ability of the proposed CD algorithm; 3) A new HSI-CD data set is designed for objective comparison of different methods. Experimental results on real hyperspectral data sets demonstrate the proposed method outperforms most of the state-of-the-arts.
\end{abstract}



%
\IEEEpeerreviewmaketitle

\ifCLASSOPTIONcompsoc
\IEEEraisesectionheading{\section{introduction}\label{sec:introduction}}
\else
\section{introduction}
\label{sec:introduction}
\fi

%
%
%
%
\IEEEPARstart{T}HE comprehensive cognition of the global change is a critical task for providing timely and accurate information in the Earth's evolution. Change detection (CD), as an important application of identifying differences of multitemporal remote sensing images, finds great use in land cover mapping \cite{DBLP:journals/tgrs/BruzzoneS97}, natural disaster monitoring \cite{Koltunov2007Early}, resource exploration \cite{Tuomas1998An} and so forth. The complete CD process is mainly divided into three steps: (1) Image preprocessing, where multitemporal images are spatially and radiometrically processed with geometric correction, radiometric correction and noise reduction; (2) Change difference image generation, where change difference images are obtained to contrast changed and unchanged regions of multitemporal images; and (3) Evaluation, where a certain measure is used to assess the performance. 

Hyperspectral image (HSI) with high spectral resolution can provide richer spectral information than other remote sensing images, such as synthetic aperture radar (SAR) images \cite{L2017Change} and multispectral images (MSIs) \cite{Usha2017Unsupervised}. Thus, HSIs have the potential to identify finer changes \cite{tewkesbury2015critical}, with the detailed composition of different objects being reflected. Though HSI-CD methods have been developed, it does not mean that the hyperspectral image change detection (HSI-CD) problem can be readily solved, for CD is a complicated process and can be affected by many factors. The main challenges for HSI-CD are summarized as follows.

1)	{\bfseries Mixed pixel problem}. In actual HSI-CD, mixed pixels commonly exist. A mixed pixel is a mixture of more than one distinct substance \cite{Keshava2002Spectral}. If one mixed pixel is roughly categorized into a certain kind of substance, it will not be expressed accurately. In general, mixed pixels may limit the improvement of hyperspectral task if not being carefully handled. 

2)	{\bfseries High dimensionality}. The high dimension of hyperspectral data makes it difficult to handle in some CD algorithms. In spite of the fact that feature extraction or band selection \cite{Qi2018Optimal} is utilized to reduce the dimensionality, the detailed information may be lost to some extent.

3)	{\bfseries Limited data sets}. The data sets for HSI-CD are relatively limited, for constructing a ground-truth map that reflects real change information of ground objects requires considerable time and effort. Both field work and manual labeling are time consuming and expensive. 

Aiming at the aforementioned three challenges, our motivations are described in detail from two aspects. First, an in-depth analysis of internal pixel is necessary, but many related research are case studies on limited problems about HSI-CD. Spectral unmixing is a process to decompose a mixed pixel into a collection of endmembers and the corresponding abundance maps. An abundance map \cite {Keshava2002Spectral} indicates the proportion of a corresponding endmember present in the pixel, containing much detailed subpixel composition. By considering the useful information, unmixing in HSI-CD not only enhances the performance \cite{hsieh2006subpixel}, but also provides easily interpretable information for the nature of changes \cite{Alp2015Sparse}. Though spectral unmixing has been widely studied, research on HSI-CD by unmixing is still in an early stage. Existing studies in the literature are case studies on certain problems and mainly focus on linear or nonlinear unmixing without combining them in HSI-CD.

Second, deep learning can better handle the problem of high dimension and exploit effective features, but the features are not fully mined in many HSI-CD methods. By learning which part or what kind of combination of multi-source data is critical, deep learning offers flexible approaches to deal with HSI. Generally, a change difference image can be obtained from the two preprocessed images with the trained deep neural network. However, many HSI-CD methods based on deep learning do not fully exploit the feature contained between the spectra of corresponding pixels in multitemporal images. These methods mainly analyze the changes of pixels by 1-D spectral vector. The features between the spectra of corresponding pixels contain richer information, which is often neglected.

Based on the aforementioned two motivations, this paper develops a new general framework, namely GETNET for HSI-CD task. The main contributions are summarized as follows.
\begin{itemize}
\item[(1)] A simple yet effective mixed-affinity matrix is proposed to better mine the changed patterns between two corresponding spectral vectors on a spatial pixel. Different from previous methods, mixed-affinity matrix converts two 1-D pixel vectors to a 2-D matrix for CD task, which provides more abundant cross-channel gradient information. Moreover, mixed-affinity matrix can effectively deal with multi-source data simultaneously and make it possible to learn representative features between the spectra in the GETNET.
\item[(2)]	A novel end-to-end framework named GETNET based on 2-D CNN is developed to improve the generalization ability of HSI-CD algorithm. To the best of our knowledge, GETNET is the first one to fuse HSI and abundance maps together to handle changes caused by different complicated factors. The proposed approach outperforms (or is competitive to) other methods across all the four data sets without parameter adjustment. Moreover, this approach offers more potential than other methods on some special data sets given that unmixing information is properly used.
\item[(3)]	A new HSI-CD data set named ``river'' in Fig. \ref{newdataset} is designed for objective comparison of different methods, and the corresponding ground-truth map is shown in Fig. \ref{river} (g). The changed information on this data set is relatively complicated and experimental results show that this data set we create is reliable and robust for quantitative assessment of CD task.
\end{itemize}

The remainder of this paper is organized as follows. Section \ref{related work} reviews the related work. Section \ref{OurApproach} describes the detail of the proposed method. Section \ref{Experiments} evaluates the performances of different CD approaches. Finally, we conclude this paper in Section \ref{Conclusion}.
\ifCLASSOPTIONcompsoc
\IEEEraisesectionheading
\else
\section{related work}
\label{related work}
\fi

\begin{figure*}
	\centering
	\quad
	\quad
	\includegraphics[width=0.75\textwidth]{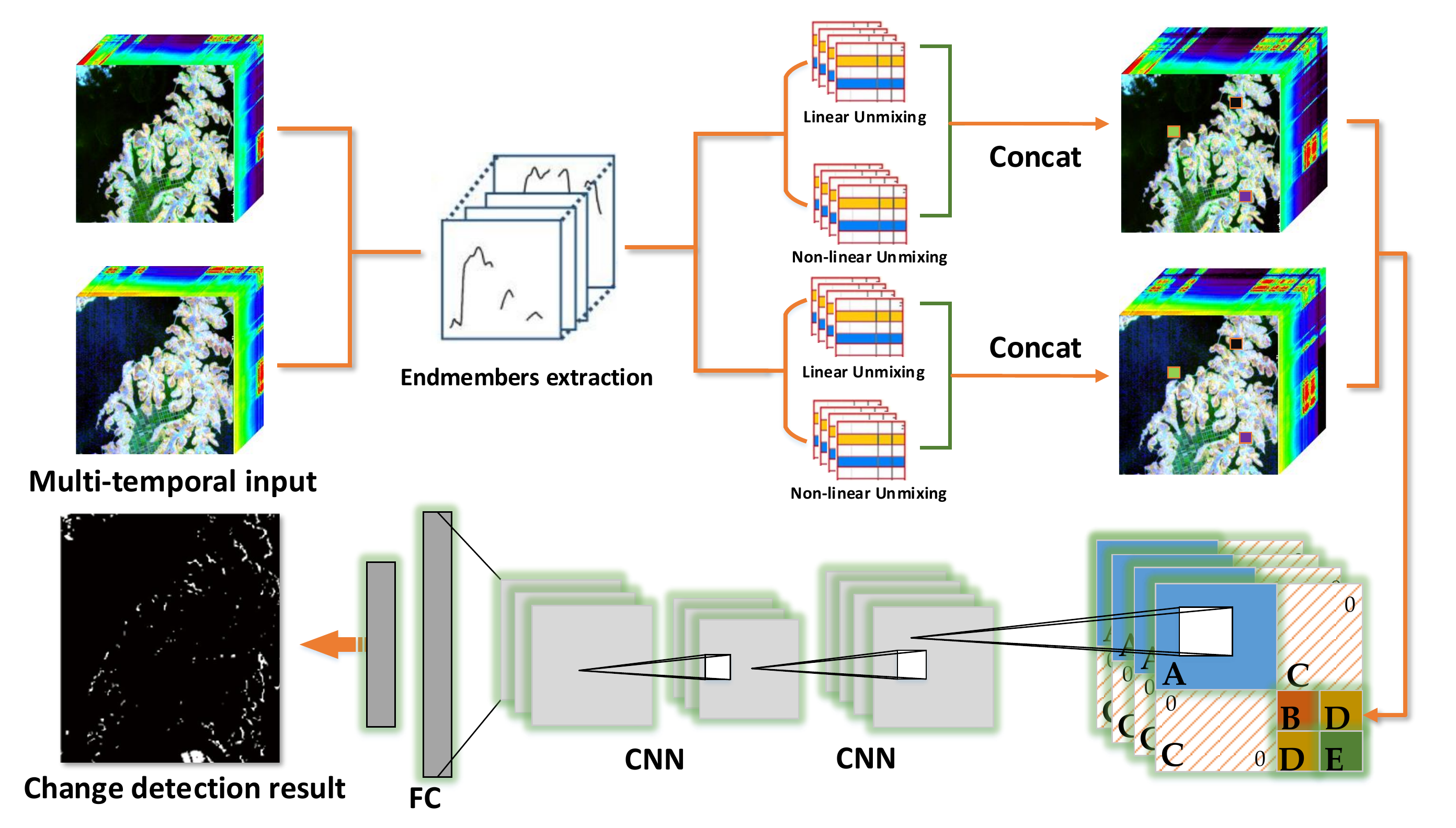}
	\caption {Overview of the GETNET CD method for multitemporal HSIs.}
	\label{fig:2} 
\end{figure*}

The most common and simple methods of CD are image differencing and image ratioing, mainly designed for single-band images. These two methods are easy to implement but their applicable scope is limited. When handling multi-band remote sensing images, classical CD methods fall primarily into four categories.

1) \emph{Image arithmetical operation}. The most common approach is change vector analysis (CVA) \cite{carvalho2011new}, which generates magnitude and direction of change by spectral vector subtraction. Besides, there are many comparative analyses based on CVA, such as ICVA, MCVA and CVAPS \cite{Bovolo2007A}. Typically, based on the polar CVA, ${\rm C^{2}VA}$ \cite{Bovolo2012A} describes the CD problem in a magnitude-direction 2-D representation produced by a loss compression procedure. In addition, Chen \emph{et al}. \cite{Chen2015Change} apply tensor algebra to CD and propose 4-D Higher Order Singular Value Decomposition (4D-HOSVD) to capture comprehensive changed features.

2) \emph{Image transformation}. Methods based on image transformation convert multispectral images into a specific feature space to emphasize changed pixels and suppress unchanged ones. As a linear transformation technique, Principal Component Analysis (PCA) \cite{Zhang2016Feature} decorrelates images \cite{Celik2009Unsupervised}. However, PCA heavily relies on the statistical property of images and is especially vulnerable to unbalanced data \cite{Deng2008PCA}. Based on canonical correlation analysis, multivariate alteration detection (MAD) \cite{Nielsen1997Multivariate} is an unsupervised method to analyze multi-temporal images CD. Nevertheless, MAD is not applicable to affine transformation. Then, iteratively reweighted multivariate alteration detection (IR-MAD) \cite{Nielsen2007The} extends MAD by applying different weights to observations. However, IR-MAD still cannot utilize the notable relationship between multitemporal bands. The subspace-based change detection (SCD) method \cite{Wu2013A} takes the observed pixel of one HSI as target, and establishes the background subspace by utilizing the corresponding pixel in the other image.

3) \emph{Image classification}. These methods are supervised schemes and have prior knowledge for the training of a classifier. Among them, post-classification comparison (PCC) \cite{Bovolo2012A} takes the CD as an image pair classification task, where the two classified images are compared pixel by pixel. The pixels fall into different categories are considered as changed regions. This is also the case of classification of differential features \cite {Nemmour2006Multiple} and compound classification \cite{Bruzzone1997An}.

4) \emph{Other advanced methods}. Research on deep learning has found significant use in CD of multi-band images. For example, Zhang \emph{et al}. \cite{Zhang2016Feature} develop a deep belief network (DBN) based feature extraction method, which concatenates a pair of multitemporal vectors to output the deep representation used for CD. By reshaping pixels to vectors, Zhao \emph{et al}. \cite{zhao2014deep} present a restricted Boltzmann machine (RBM) based deep learning methods to learn representation for image analysis. In addition, more new approaches have been put forward to solve the problems of HSI-CD in the recent studies. Liu \emph{et al}. \cite{Liu2015Hierarchical} describe a hierarchical CD approach, aiming at distinguishing all the possible change types by considering discriminable spectral behaviors. Based on statistical analyses of spectral profiles, Han \emph{et al}. \cite{Han2017An} adopt an unsupervised CD algorithm for denoising without dimensionality reduction.

The aforementioned four categories can be applied to HSIs, but the accuracy may be improved if mixed pixels are considered to handle pixel level change. In order to address mixed pixels in HSIs, unmixing is necessary. HSI-CD by unmixing has obvious advantages of providing subpixel level information. However, unmixing in HSI-CD is very challenging and has not been extensively studied so far. The work in \cite{ErturkP15} obtains change difference image by calculating changes between the corresponding abundance maps of multitemporal images. In addition to spectral information, superpixels are utilized to integrate spatial information in the unmixing process \cite{Alp2016Unmixing}. These two approaches detect changes in the abundance of multitemporal images for each endmember simply by difference operation, limited to certain problems. Liu \emph{et al}. \cite{Liu2015Multitemporal} analyzes abundance maps of changed and unchanged endmembers and their contribution to each pixel. Besides, sparse unmixing is firstly explored for HSI-CD by Ert\"{u}rk \emph{et al}. \cite{Alp2015Sparse} with spectral libraries. Many studies about unmixing in change analysis are limited to case studies and need to be further explored.      

\section{METHODOLOGY}
\label{OurApproach}
The overview of the proposed CD method for multitemporal HSIs is shown in Fig. \ref{fig:2}. 
First, spectral unmixing algorithms are used to deal with two preprocessed HSIs. Then the abundance maps obtained by linear and nonlinear unmixing interact with the HSIs and form mixed-affinity matrices. Next mixed-affinity matrices are processed by the GETNET. In the end, the final CD result is the output of the deep neural network.

In this section, we introduce our method from three aspects: hybrid-unmixing for subpixel level mining, mixed-affinity matrix generation for information fusion and GETNET with ``separation and combination''.

\subsection{\textbf{Hybrid-unmixing for Subpixel Level Mining}}
\label{gcdm}

In order to obtain subpixel level information and improve generalization ability, spectral unmixing is utilized in the proposed method. A complete spectral unmixing process contains endmember extraction and abundance estimation. In previous studies, various endmember extraction algorithms(EEAs) are developed and any EEA is applicable to the proposed method. In this work, automatic target generation process (ATGP) \cite{Ren2010Automatic} is adopted to find potential target endmembers. After endmembers are obtained, the next step is to achieve abundance estimation, namely, abundance maps for each pixel.

Aiming at achieving abundance estimation, mixture models can be divided into two categories: linear and nonlinear. Either has advantages and is applicable to certain cases \cite{Chen2013Nonlinear}, \cite{Yu2017Comparison}. Nevertheless, few works pay attention to combining them in HSI-CD. In order to utilize both advantages, our study focuses on hybrid-unmixing, namely, not only linear but also nonlinear mixture model.

The linear mixture model assumes that each pixel is a linear combination of endmembers and the reflectance \pmb{${r}$} of one pixel can be expressed as
\begin{center}
\begin{flalign}
		\label{spatial}
		\boldsymbol{r}=\sum\limits_{i = 1}^m {w_{l_i}} \boldsymbol{{x_i}} +\varepsilon= \boldsymbol{X{w_l}}+\varepsilon, \\
		\textbf{s.t.} \sum\limits_{i = 1}^m {w_{l_i}}  = 1,\\
		 0 \le w_{l_i} \le 1, i = 1,2,3,...,m ,
\end{flalign}
\end{center}
where \pmb{${r}$} represents a ${b \times 1}$ vector; ${b}$ is the number of HSI bands; ${m}$ is the number of endmembers; \pmb{${X}$} is a ${b \times m}$ matrix of endmembers; \pmb{${x_i}$} is the ${i}$th column of \pmb{${X}$}; \pmb{{${w_l}$}} is an ${m \times 1}$ vector of linear abundance fraction; ${w_{l_i}}$ denotes the linear proportion of the ${i}$th endmember; ${\varepsilon}$ is a ${b \times 1}$ vector of noise caused by the sensor and modelling errors. Eqs. (2) and (3) are two constraints of abundances non-negative constraint (ANC) and abundance sum-to-one constraint (ASC), respectively. To estimate the parameters, the fully constrained least squares (FCLS) \cite{Heinz2002Fully} method is adopted. The linear abundance map \pmb{{${w_l }$}} is the key information used in the proposed method.

The nonlinear mixture model can be formulated as
\begin{center}
\begin{flalign}
	\boldsymbol{r} = \sum\limits_{i = 1}^m {{{w_n}_i} \boldsymbol{{x_i}} + \sum\limits_{i = 1}^m\sum\limits_{j = 1}^m {{a_{ij}} \boldsymbol{{x_i}} \odot \boldsymbol{{x_j}}} }+\varepsilon,\\
	{\textbf{s.t.} \forall i \ge j:{a_{ij}} = 0},\\
	\forall i < j:{a_{ij}} = {{w_n}_i}{{w_n}_j},\\
	\sum\limits_{i = 1}^m {{{w_n}_i} = 1,{{w_n}_i} \ge 0},
\end{flalign}
\end{center}
where ${\odot}$ is the point-wise multiplication operator; ${a_{ij}}$ represents the bilinear parameter to model the nonlinearity of HSIs; ${w_{n_i}}$ denotes the nonlinear proportion of the ${i}$th endmember; Other parameters have the same meaning in Eq. (1). In this work, the bilinear-Fan model (BFM) \cite{Yu2017Comparison} is employed to obtain the parameters and the nonlinear abundance map \pmb{{${w_n}$}} is achieved. It is worth noting that both inputs in one data set share the same endmembers. Specifically, all the endmembers used in FCLS and BFM algorithms are the same and generated by the ATGP algorithm. By applying simple and effective methods, subpixel level information is explored fully to the next step. As linear and nonlinear mixture models are appropriate for different cases, combining both methods in a unified framework can improve the generalization ability.

\begin{figure}
	\centering
	\includegraphics[width=0.30\textwidth]{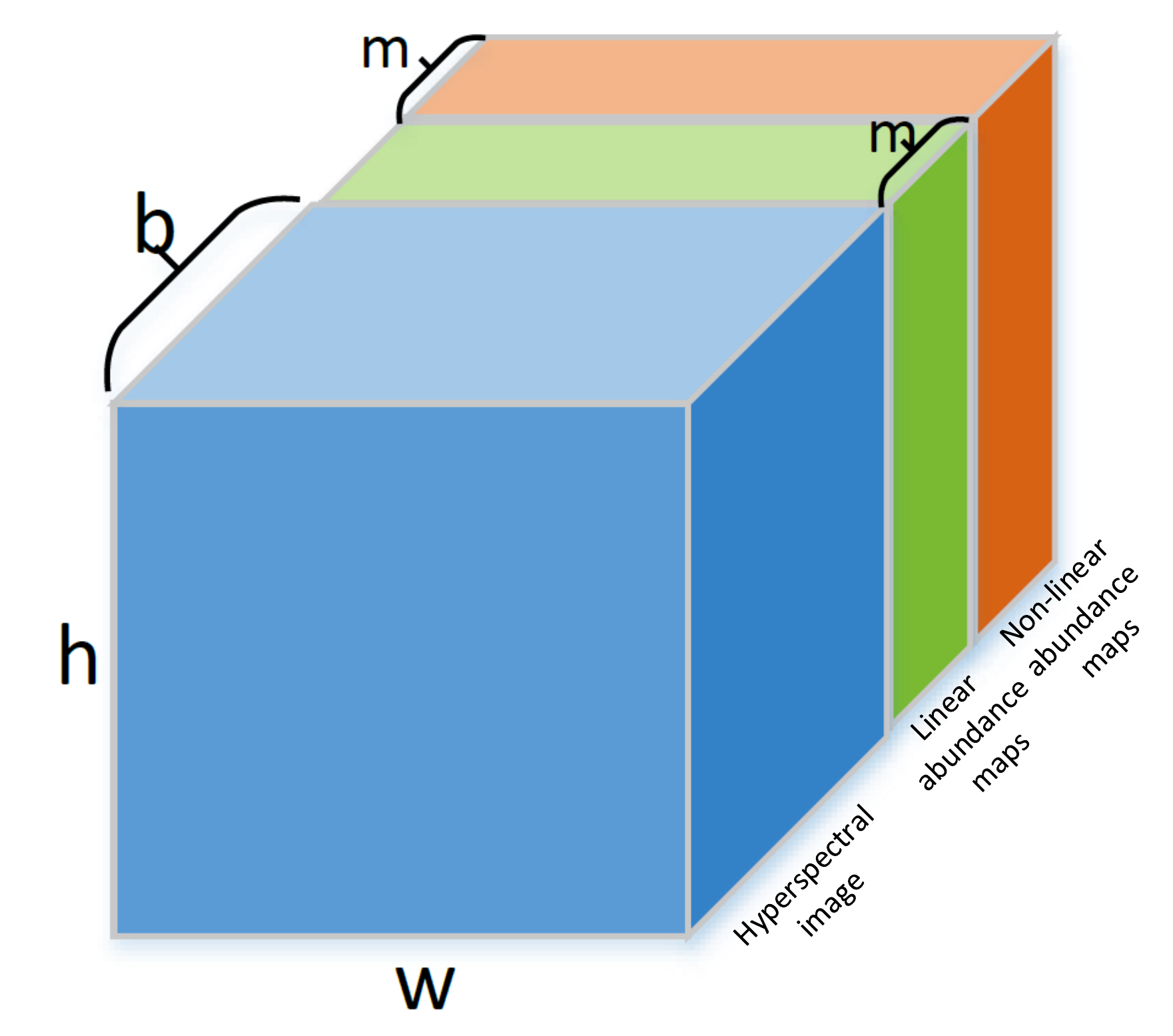}
	\caption{The mixed data cube of HSI and unmixing abundance maps.}
	\label{fig:inputs}
\end{figure}

The size of HSI is ${h \times w \times b}$. For convenience, we stack all the \pmb{{${w_l}$}} (the number of \pmb{{${w_l}$}} is ${h \times w}$) into a 3-D abundance data cube with the size of ${h\times w \times m}$. \pmb{{${w_n}$}} is processed on the same way of \pmb{{${w_l}$}}. Then, the next step is to integrate the two types of abundance maps together with the hyperspectral data. In other words, three kinds of data are stacked neatly in sequence in the direction of HSI spectral domain, as shown in Fig. \ref{fig:inputs}. The final multi-source data cube, with the size of ${w \times h \times (b+2m)}$, contains almost all the HSI information for the subsequent analysis.
 
\subsection{\textbf{Mixed-affinity Matrix Generation for Information Fusion}}
\label{track}

\begin{figure}
	\centering
	\includegraphics[width=0.30\textwidth]{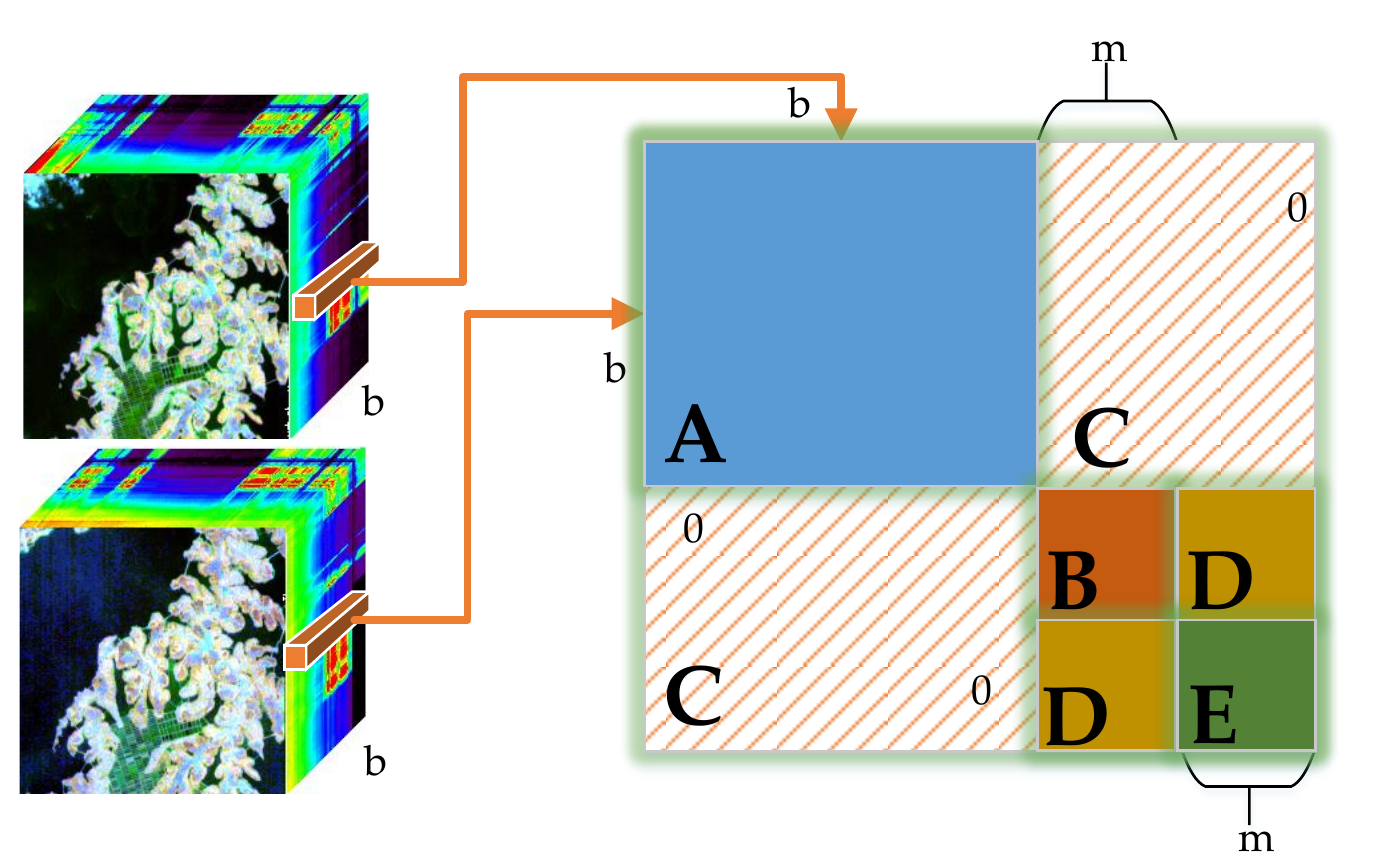}
	\caption{Mixed-affinity matrix of one pixel.}
	\label{fig:aff}
\end{figure}

Different from the previous deep learning methods, this work proposes a novel mixed-affinity matrix for 2-D CNN algorithm. Assume there are ${b}$ bands to represent one pixel in HSI. Moreover, there are ${m}$ different endmembers and each endmember has a complete linear and nonlinear abundance maps (the size of complete abundance map is ${h \times w}$). As described in Section. \ref{gcdm}, both pixel and subpixel level information of the integrated multi-source data cube is obtained. Hence, the similarity between different bands of the corresponding pixels is a kind of signature to describe this pixel, i.e., a certain band in one pixel interacts with other ${b}$ bands in the corresponding pixel respectively. Meanwhile, the similarity between different abundance maps is also another signature to indicate subpixel representation. Then we define mixed-affinity matrix, which contains both pixel and subpixel level signatures to represent each pixel. The mixed-affinity matrix is constructed as
\begin{equation}
\label{affinity}
K{_{ij}} = 1 - ({r_{1_{ij}}} - {r_{2_{ij}}})/{r_{2_{ij}}}, i,j = 1,2,3,...,N
\end{equation}
where $K{_{ij}}$ is a 2-D mixed-affinity matrix of ${n \times n}$ and it represents the relationship between corresponding pixels of different wavelengths and abundance maps; ${N={h \times w }}$; ${n=b+2m}$; ${r_{1_{ij}}}$ is a ${n \times 1}$ vector of a pixel at ${time_1}$ in Fig. \ref{fig:inputs}; ${r_2}$ is another ${n \times 1}$ vector of the corresponding pixel at ${time_2}$. Mixed-affinity matrix is a 2-D matrix of ${n \times n}$ and it represents the relationship between corresponding pixels of different wavelengths and abundance maps, i.e., mixed-affinity. As illustrated in Fig. \ref{fig:aff}, mixed-affinity matrix is divided into five different parts.

In part A, a certain band at ${time_1}$ in one pixel subtracts from other ${b}$ bands in the corresponding pixel at ${time_2}$. Part A represents the pixel level difference information. In part B, linear abundance map of a certain endmember for one pixel at ${time_1}$ subtracts from the corresponding linear abundance map at ${time_2}$. Similarly, in part E, nonlinear abundance map interacts with each other by subtraction. In part D, linear abundance map subtracts from the corresponding nonlinear abundance map. Part B, D and E reveal subpixel level difference information. As the affinity between hyperspectral data and abundance maps is meaningless, each ${K_{ij}}$ in part C is set to 0. In general, the image spectral affinity and abundance affinity are distributed in the left top and right bottom corner respectively in Fig. \ref{fig:aff}.

\begin{figure}
	\centering
	\includegraphics[width=0.50\textwidth]{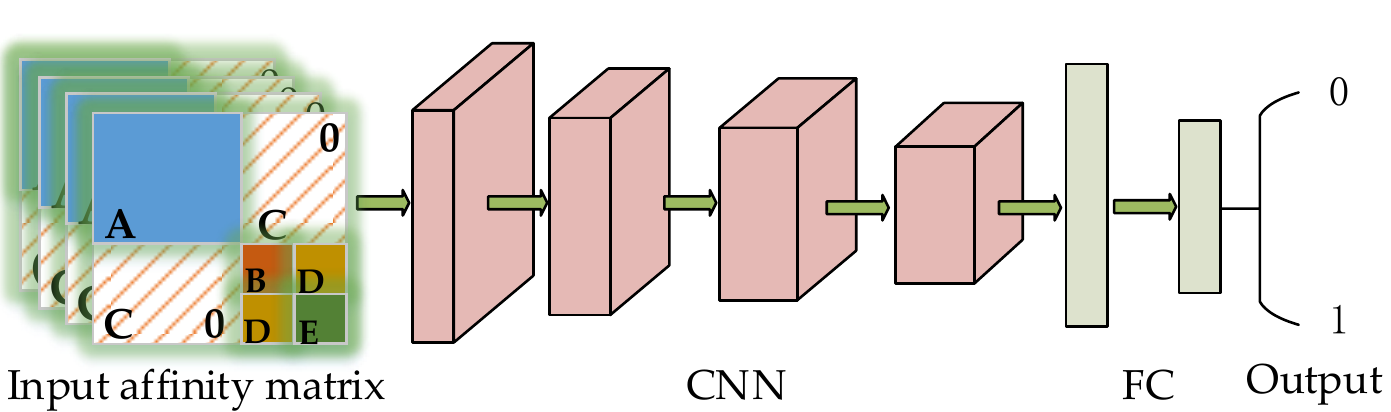}
	\caption{Architecture of the proposed CNN with mixed-affinity matrices for CD.}
	\label{fig:cnn}
\end{figure}

The greater value of $K{_{ij}}$ is, the more similar it is between the corresponding pixels. After the mixed-affinity matrices are calculated, they are input to the proposed 2-D CNN. The architecture takes the CD as a binary classification task and outputs the classification result shown in Fig. \ref{fig:cnn}. Remarkably, the similarity and difference of the corresponding pixels can be seen from the value of mixed-affinity matrix. Mixed-affinity matrix has important significance in three aspects as follows.
 
1)	Mixed-affinity matrix is an efficient way for simultaneous processing of multi-source information fusion, combining preprocessed hyperspectral data with linear and nonlinear abundance maps. In addition, pixel and subpixel level representation is achieved naturally.

2)	The differences between the two 1-D pixel vectors is mapped to a 2-D matrix, which provides more abundant cross-channel gradient information. By mining the differences among ${b}$ channels and ${2m}$ complete abundance maps, the utilization ratio of multi-source information is maximized. 

3)	The powerful learning ability of the 2-D CNN can work seamlessly with the mixed-affinity matrix. On the one hand, deep learning is able to learn the significant features for CD task. On the other hand, it can integrate information with local context according to different parts in mixed-affinity matrix.

\subsection{\textbf{GETNET with ``Separation and Combination"}}
\label{netdesign} 
CD can be seen as binary classification of changed and unchanged pixels. The proposed classification model on mixed-affinity matrix is different from a traditional CNN classification task. In common image classification task, better performance is usually obtained by a deeper CNN model. This model commonly exploits residual layers or inception modules to learn powerful representations of images. Considering the specific CD task, we design a general end-to-end 2-D CNN architecture for the vector pair classification problem. The end-to-end architecture can help ensure high accuracy of the whole system by optimizing it at all layers and generate the change difference image without stage-wise optimization or error accumulation problem. Then, the designed network is introduced as follows.

\subsubsection{\textbf{``Separation and Combination"}}
The proposed algorithm involves convolution operators on the mixed-affinity matrices. The structure of mixed-affinity matrix is illustrated in Fig. \ref{fig:aff}, where the image spectral affinity and abundance affinity are distributed in the left top and right bottom corner respectively. Traditional convolution operation shares kernel weights on the whole input map. However, it is not suitable for the mixed-affinity matrix at the early convolution layer, since spectral feature has different nature compared with the abundance map. In this work, two different convolution kernels are operated on the mixed-affinity matrix. One is for the image spectral affinity in the left top, i.e., part A, and the other is for the abundance affinity in the right bottom corner, i.e., part B, D and E. Because the value of part C is 0, any convolution kernel is applicable here. Then we adopt the locally sharing weights convolution \cite{huang2012learning} for these parts.

After locally sharing convolution layers and max pooling layers on the mixed-affinity matrices, two kinds of higher level features of different parts, i.e., spectral feature and abundance feature, are learned automatically by the proposed network. The next step is to classify these features by the fully connected layers, which work as a classifier. Additionally, considering the truth that different data sets are sensitive to different types of features, the fully connected layers are employed to fuse these features to improve the generalization ability.

\subsubsection{\textbf{Training with Pre-detection}}
The overall architecture of the proposed network is shown in Fig. \ref{fig:cnn}, and the detailed parameters are presented in Table \ref{tab1}, where the “LS” stands for locally sharing weights convolution \cite{huang2012learning}. In this work, HSI-CD is viewed as a binary classification task of changed and unchanged pixels. Mixed-affinity matrices are input into the 2-D CNN and the outputs of network are the class labels of pixels. A change difference image can be obtained from the two HSIs with the trained network. However, the ground-truth map is used to evaluate the different CD methods but not for training in HSI analysis. Thus, we need to generate pseudo data sets to train it. Any HSI-CD method can be used to generate labeled samples. This research adopts the conventional method CVA with high confidence threshold to generate pseudo training sets with labels. Pixels which have high possibility of being classified correctly are selected as final labeled samples to train the network. The positive samples (changed samples) account for about 10\% of the changed pixels. The number of negative samples (unchanged samples) is twice of the positive samples, namely that the ratio of positive to negative samples is 1 : 2.

The number of training samples has a significant impact on the performance of classification. If the positive and negative training samples are unbalanced, the network may be biased to the class with more samples. To avoid this, the ratio between the positive and negative samples is set to 1:1 at first. However, more training samples is desired to train a deep network without pre-trained models. According to the pre-detection method (CVA) output, the ratio between the changed and unchanged samples is usually far less than 1:1. Thus more unchanged samples are needed to train the network. Then the ratio of positive to negative samples is set to 1 : 2 in this model.

\begin{table}[]
	\centering
	\caption{Architecture details for GETNET.}
	\label{tab1}
	\scalebox{0.85}{
	\begin{tabular}{|c|c|c|c|}
		\hline
		Layers   & Type                                                                            & Channels & Kernel Size \\ \hline
		LSConv1  & \begin{tabular}[c]{@{}c@{}}LSConvolution + BN\\ Activation(tanh)\end{tabular}   & 32       & 5 x 5       \\ \hline
		MaxPool1 & MaxPooling                                                                      & -        & 2 x 2       \\ \hline
		LSConv2  & \begin{tabular}[c]{@{}c@{}}LSConvolution + BN\\ Activation(tanh)\end{tabular}   & 64       & 3 x 3       \\ \hline
		MaxPool2 & MaxPooling                                                                      & -        & 2 x 2       \\ \hline
		LSConv3  & \begin{tabular}[c]{@{}c@{}}LSConvolution + BN\\ Activation(tanh)\end{tabular}   & 128      & 3 x 3       \\ \hline
		MaxPool3 & MaxPooling                                                                      & -        & 2 x 2       \\ \hline
		LSConv4  & \begin{tabular}[c]{@{}c@{}}LSConvolution + BN\\ Activation(tanh)\end{tabular}   & 96       & 1 x 1       \\ \hline
		MaxPool4 & MaxPooling                                                                      & -        & 2 x 2       \\ \hline
		FC1      & \begin{tabular}[c]{@{}c@{}}Fully Connected + BN\\ Activation(tanh)\end{tabular} & 512      & -           \\ \hline
		FC2      & Fully Connected                                                                 & 2        & -           \\ \hline
	\end{tabular}}
\end{table}

\section{Experiments}
\label{Experiments}
In this section, extensive experiments are conducted using four real HSI data sets. Firstly, we introduce the existing and newly constructed data sets used in the experiments. Then evaluation measures are described for HSI-CD methods. In the end, the experimental performances of the proposed method and other state-of-the-arts are analyzed in detail.

\subsection{\textbf{Datasets Description}}
\label{taun}

The most important reason for the lack of HSI-CD data sets is that it is difficult to construct a ground-truth map. Ground-truth map is a change difference image reflecting the real change information of ground objects, composed of changed and unchanged class. Each data set contains three images, two real HSIs which are taken at the same position from different times, and a ground-truth map. Generally, the structure of ground-truth map mainly depends on the on-the-spot investigation and some algorithms with high accuracy. For uncertain pixels, it is necessary to query their spectral values for further judgment. Constructing precise ground-truth map is vital for its quality directly affects accuracy assessment.

\begin{figure}
	\centering
	\includegraphics[width=0.45\textwidth]{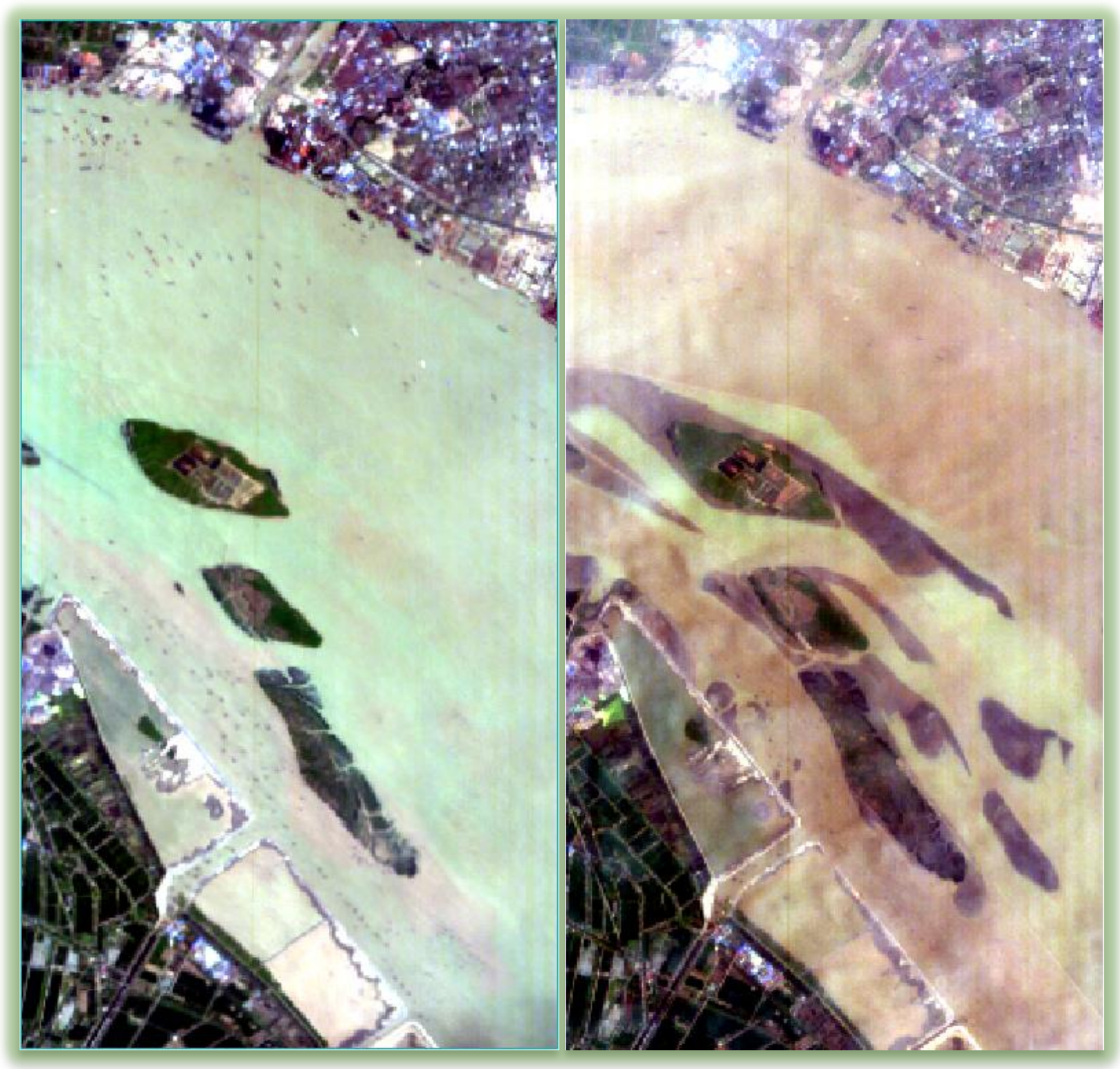}
	\caption{The newly constructed data set of HSIs at two different time: (a) The river imagery on May 3, 2013, (b) The river imagery on December 31, 2013.}
	\label{newdataset}
\end{figure}

In this work, all the HSI data sets used in the experiments are selected from Earth Observing-1 (EO-1) Hyperion images. The EO-1 Hyperion covers the 0.4-2.5 microns’ spectral range, with 242 spectral bands. Moreover, it provides a spectral resolution of 10 nm approximately, as well as a spatial resolution of 30 m. Though there are 242 spectral bands in Hyperion, the HSI quality can be harshly affected by the atmosphere. Typically, some of spectral bands are noisy with low signal-to-noise ratio (SNR). Therefore, according to the existing research and ENVI analysis, bands with high SNR are selected in the following analysis. The experiments are carried out on the data sets after noise elimination and the ground-truth maps of existing data sets come from \cite{Yuan2015Semi}.

\subsubsection{\textbf{Existing Data Sets}}
\begin{figure}
	\centering
	\includegraphics[width=0.45\textwidth]{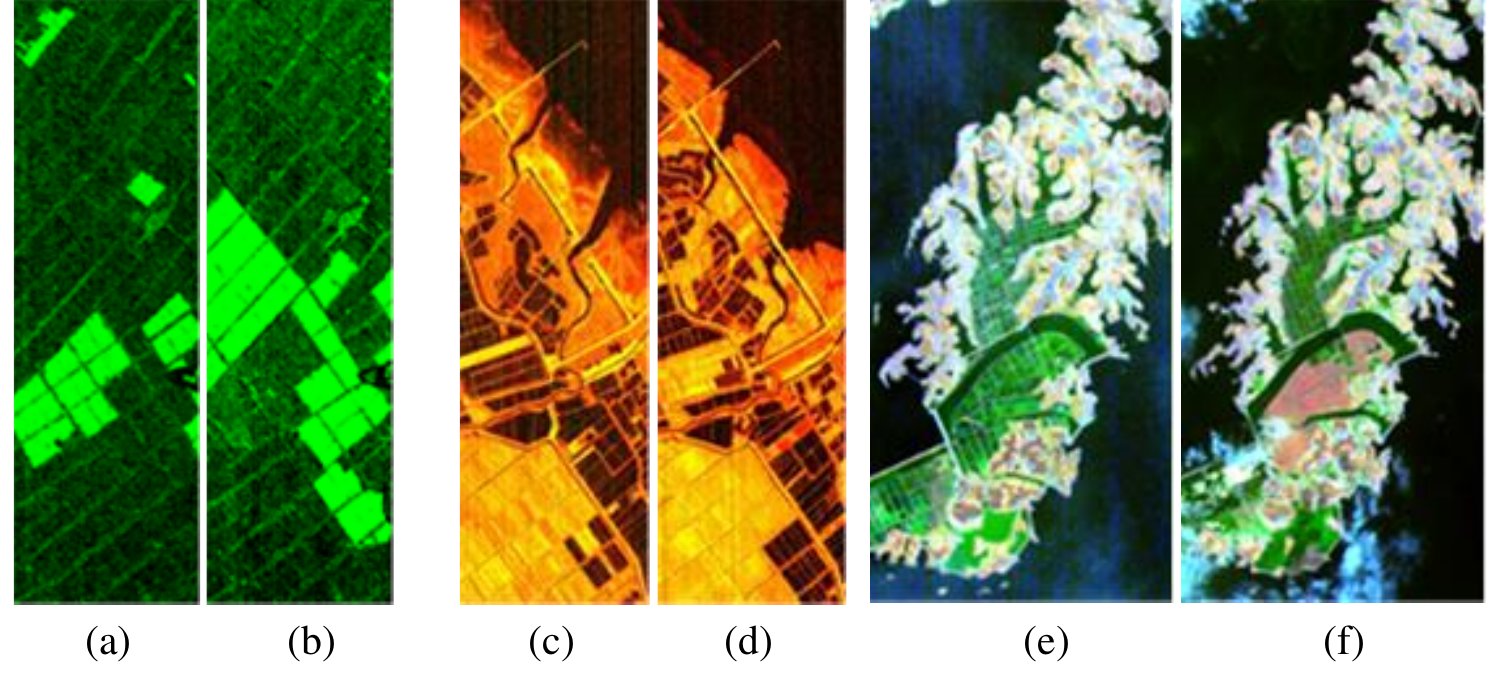}
	\caption{The existing experimental data sets in our paper: (a) The farmland imagery on May 3, 2006, (b) The farmland imagery on April 23, 2007, (c) The countryside imagery on November 3, 2007, (d) The countryside imagery on November 28, 2008, (e) The Poyang lake imagery on July 27, 2002, (f) The Poyang lake imagery on July 16, 2004.}
	\label{existingdatasets}
\end{figure}

There are some available hyperspectral data sets for HSI-CD methods. The first data set ``farmland" is shown in Fig. \ref{existingdatasets} (a) and (b). It covers farmland near the city of Yancheng, Jiangsu province, China, with a size of ${450 \times 140}$ pixels. The two HSIs were acquired on May 3, 2006, and April 23, 2007, respectively. There are 155 bands selected for CD after noise elimination. Visually, the main change on this data is the size of farmland. In this experiment, about 20.95\% pixels of the whole image are selected as labeled samples, with 4400 positive samples and 8800 negative samples. There are more samples selected than other data sets because the proportion of changed pixels is higher.

The second data set ``countryside" is shown in Fig. \ref{existingdatasets} (c) and (d), which covers countryside near the city of Nantong, Jiangsu province, China, having a size of ${633 \times 201}$ pixels. The multitemporal HSIs were obtained on November 3, 2007 and November 28, 2008, respectively. 166 bands are selected for CD after noise elimination. Visually, the main change on this data is the size of rural areas. In this experiment, about 7.55\% pixels are selected as labeled samples, with 3200 positive samples and 6400 negative samples.

The third data set ``Poyang lake" is shown in Fig. \ref{existingdatasets} (e) and (f). It covers the province of Jiangxi, China, with ${394 \times 200}$ pixels. The two HSIs were obtained on July 27, 2002 and July 16, 2004, respectively. 158 bands are available for CD after noise elimination. Visually, the main change type on this data is the land change. In this experiment, about 2.17\% pixels are selected, where 570 positive samples and 1140 negative samples are taken as labeled data.

\subsubsection{\textbf{Newly Constructed Data Set}}The newly constructed data set ``river" \footnote[1]{http://crabwq.github.io} is shown in Fig. \ref{newdataset}. The two HSIs were acquired at May 3, 2013, and December 31, 2013, respectively in Jiangsu province, China. This data set has a size of ${463 \times 241}$ pixels, with 198 bands available after noisy band removal. The main change type on this data is disappearance of substance in river. In this experiment, about 3.37\% pixels are selected as labeled samples, with 1250 positive samples and 2500 negative samples.

\begin{table}[]
	\centering
	\caption{confusion matrix}
	\label{confusion-matrix}
	\begin{tabular}{|l|l|l|l|l|}
		\hline
		\multicolumn{2}{|l|}{\multirow{2}{*}{\begin{tabular}[c]{@{}l@{}}Confusion\\  Matrix\end{tabular}}} & \multicolumn{3}{l|}{Predicted} \\ \cline{3-5} 
		\multicolumn{2}{|l|}{}                                                                             &          & 1        & 0        \\ \hline
		\multicolumn{2}{|l|}{\multirow{2}{*}{Actual}}                                                      & 1        & TP       & FN       \\ \cline{3-5} 
		\multicolumn{2}{|l|}{}                                                                             & 0        & FP       & TN       \\ \hline
	\end{tabular}
\end{table}

\subsection{\textbf{Evaluation Measures}}

In order to evaluate the performance of CD methods, we compare the change difference images with ground-truth maps, in which white pixels represent changed portions and black pixels mean unchanged parts. The accuracy of result can directly reflect the reliability and practicability of the CD methods. Generally, through pixel level evaluation, this paper adopts three evaluation criteria: overall accuracy(OA), Kappa coefficient and confusion matrix. In their calculation, there are four indexes: 1) true positives (TP), i.e., the number of correctly-detected changed pixels. 2) true negatives (TN), i.e., the number of correctly-detected unchanged pixels. 3) false positives (FP), i.e., the number of false-alarm pixels, and 4) the false negatives (FN), i.e., the number of missed changed pixels. The overall accuracy (OA) is defined as

\begin{equation}
OA = \frac{{TP + TN}}{{TP + TN + FP + FN}}.
\end{equation}

Kappa coefficient reflects the agreement between a final change difference image and the ground-truth map. Compared to OA, Kappa coefficients can more objectively indicate the accuracy of the CD results. The higher value of Kappa coefficient, the better result of CD will be. Kappa coefficient is calculated as
\begin{equation}
Kappa = \frac{{OA - P}}{{1 - P}},
\end{equation}
where 
\begin{equation}
\resizebox{.85\hsize}{!}{${P = \frac{{(TP + FP)(TP + FN)}}{{{{(TP + TN + FP + FN)}^2}}} + \frac{{(FN + TN)(FP + TN)}}{{{{(TP + TN + FP + FN)}^2}}}}$}
\end{equation}

As illustrated in Table \ref{confusion-matrix}, a confusion matrix represents information about predicted and actual binary classifications (changed and unchanged classifications) conducted by CD algorithms, and it indicates how the instances are distributed in estimated and true classes.

\begin{table*}[]
	\centering
	\caption{The overall accuracy and kappa coefficients of GETNET and other different state-of-the-art change detection methods on four datasets \label{tab2}}
	\label{my-label}
	\scalebox{0.94}{
		\begin{tabular}{clccccc}
			\hline
			\multicolumn{2}{c}{Different methods}        & Index & \multicolumn{4}{c}{The experiment datasets}                           \\ \cline{4-7} 
			\multicolumn{3}{c}{}                                 & farmland        & countryside     & Poyang lake     & river           \\ \hline
			\multicolumn{2}{c}{CVA}                      & OA    & 0.9523          & 0.9825          & 0.9693          & \textbf{0.9529} \\
			\multicolumn{2}{l}{}                         & Kappa & 0.8855          & 0.9548          & 0.8092          & \textbf{0.7967} \\
			\multicolumn{2}{c}{PCA-CVA}                  & OA    & 0.9668          & 0.9276          & 0.9548          & 0.9437          \\
			\multicolumn{2}{l}{}                         & Kappa & 0.9202          & 0.8216          & 0.7259          & 0.7326          \\
			\multicolumn{2}{c}{IR-MAD}                   & OA    & 0.9604          & 0.8568          & 0.8248          & 0.8963          \\
			\multicolumn{2}{l}{}                         & Kappa & 0.9231          & 0.8423          & 0.7041          & 0.6632          \\
			\multicolumn{2}{c}{SVM}                      & OA    & 0.8420          & 0.9536          & 0.9583          & 0.9046          \\
			\multicolumn{2}{l}{}                         & Kappa & 0.6417          & 0.8767          & 0.7266          & 0.6360          \\
			\multicolumn{2}{c}{CNN}                      & OA    & 0.9347          & 0.9033          & 0.9522          & 0.9440          \\
			\multicolumn{2}{l}{}                         & Kappa & 0.8504          & 0.7547          & 0.8412          & 0.6867          \\ \hline
			\multicolumn{2}{c}{GETNET(without unmixing)} & OA    & 0.9765 ${\pm}$ 0.00332         & 0.9758 ${\pm}$ 0.00649          & 0.9760 ${\pm}$  0.00258         & 0.9497 ${\pm}$ 0.00119         \\
			\multicolumn{2}{c}{}                         & Kappa & 0.9367  ${\pm}$ 0.00417      & 0.9438 ${\pm}$ 0.00219         & 0.8903 ${\pm}$ 0.00395         & 0.7662 ${\pm}$ 0.00377         \\
			\multicolumn{2}{c}{GETNET(Ours, with unmixing)}             & OA    & \textbf{0.9783 ${\pm}$ 0.00599 } & \textbf{0.9851 ${\pm}$ 0.00126} & \textbf{0.9856 ${\pm}$ 0.00365} & 0.9514 ${\pm}$ 0.00482          \\
			\multicolumn{2}{c}{}                         & Kappa & \textbf{0.9572 ${\pm}$ 0.00495} & \textbf{0.9651 ${\pm}$ 0.00173} & \textbf{0.9113 ${\pm}$ 0.00408} & 0.7539 ${\pm}$ 0.00349         \\ \hline
		\end{tabular}
	}
\end{table*}

\subsection{Experimental results}

In this work, the changes between HSIs are at pixel level variation with no subpixel level change involved. Spectral unmixing is utilized to obtain subpixel level information, which can improve the performance of pixel level CD. In order to demonstrate the effectiveness and generality of the proposed method, we compare it with other state-of-the-art methods, including change vector analysis (CVA) \cite{Malila1980Change}, principal component analysis- change vector analysis (PCA-CVA) \cite{Baisantry2012Change}, iteratively reweighed multivariate alteration detection (IRMAD) \cite{Nielsen2007The}, support vector machines (SVM) \cite{Nemmour2006Multiple}, and patch-based convolutional neural network (CNN), which takes preprocessed HSIs as input. The principle for choosing these algorithms is due to their variety and popularity. Specifically, they employ different techniques to detect changes, such as image arithmetical operation, image transformation and deep learning.

\begin{figure}
	\centering
	\includegraphics[width=0.45\textwidth]{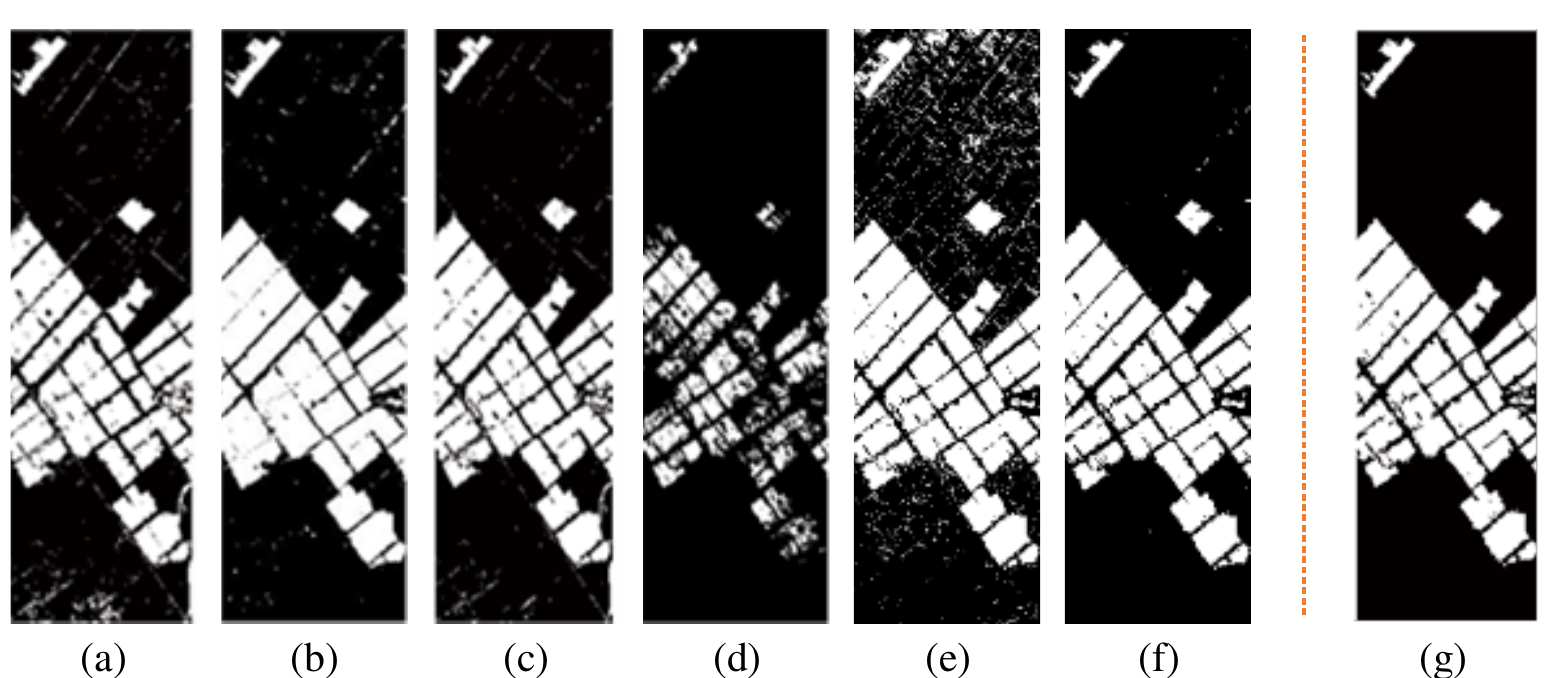}
	\caption{The change difference images of different methods on the data set of farmland, (a) the change difference image of CVA, (b) PCA-CVA, (c) IR-MAD, (d) SVM, (e) CNN, (f) GETNET, and (g) ground-truth map.}
	\label{farmland}
\end{figure}

\begin{figure}
	\centering
	\includegraphics[width=0.45\textwidth]{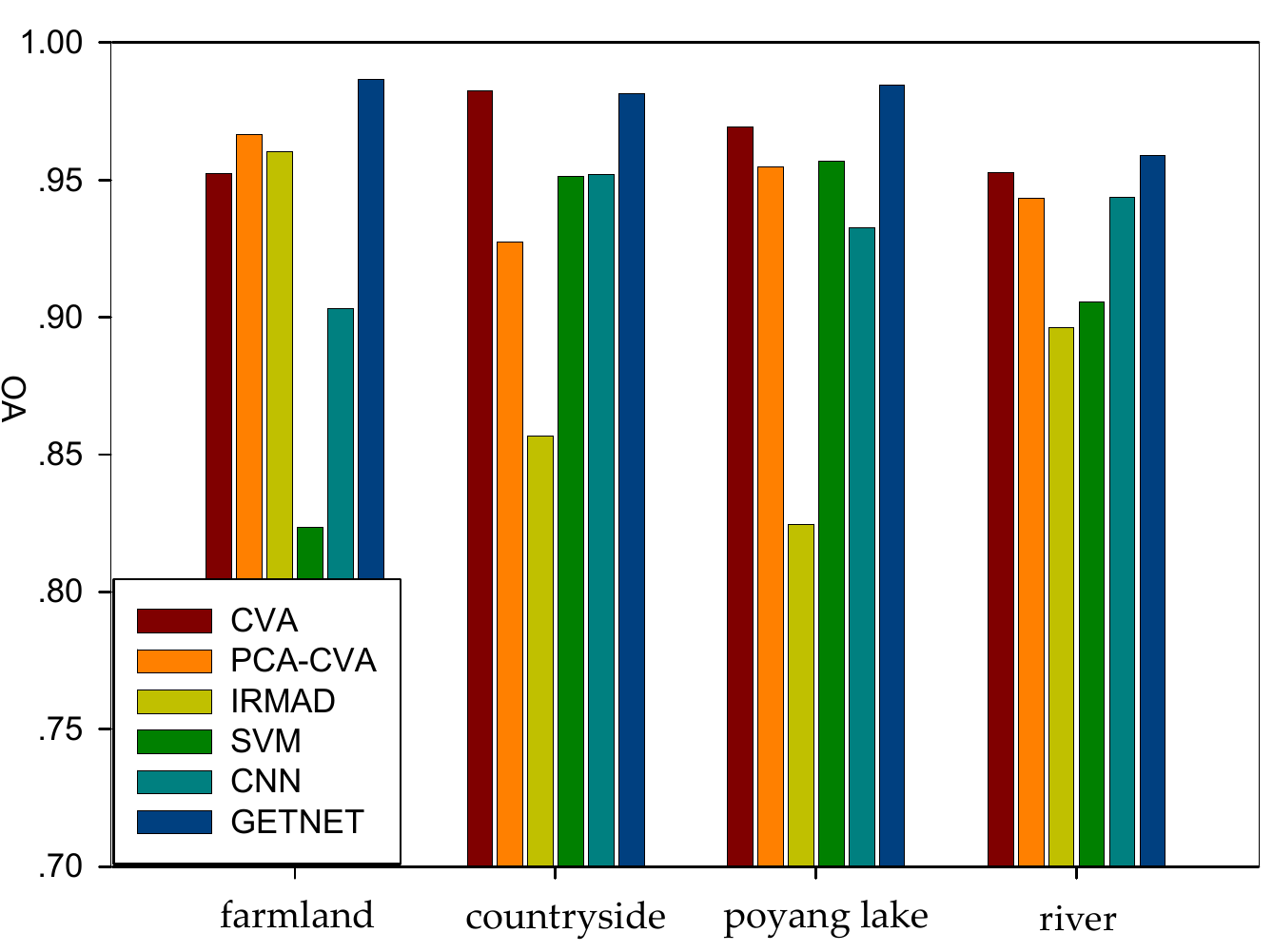}
	\caption{The OA comparisons on four data sets in different methods.}
	\label{bar_OA}
\end{figure}

\begin{figure*}
	\centering
	\includegraphics[width=1.0\textwidth]{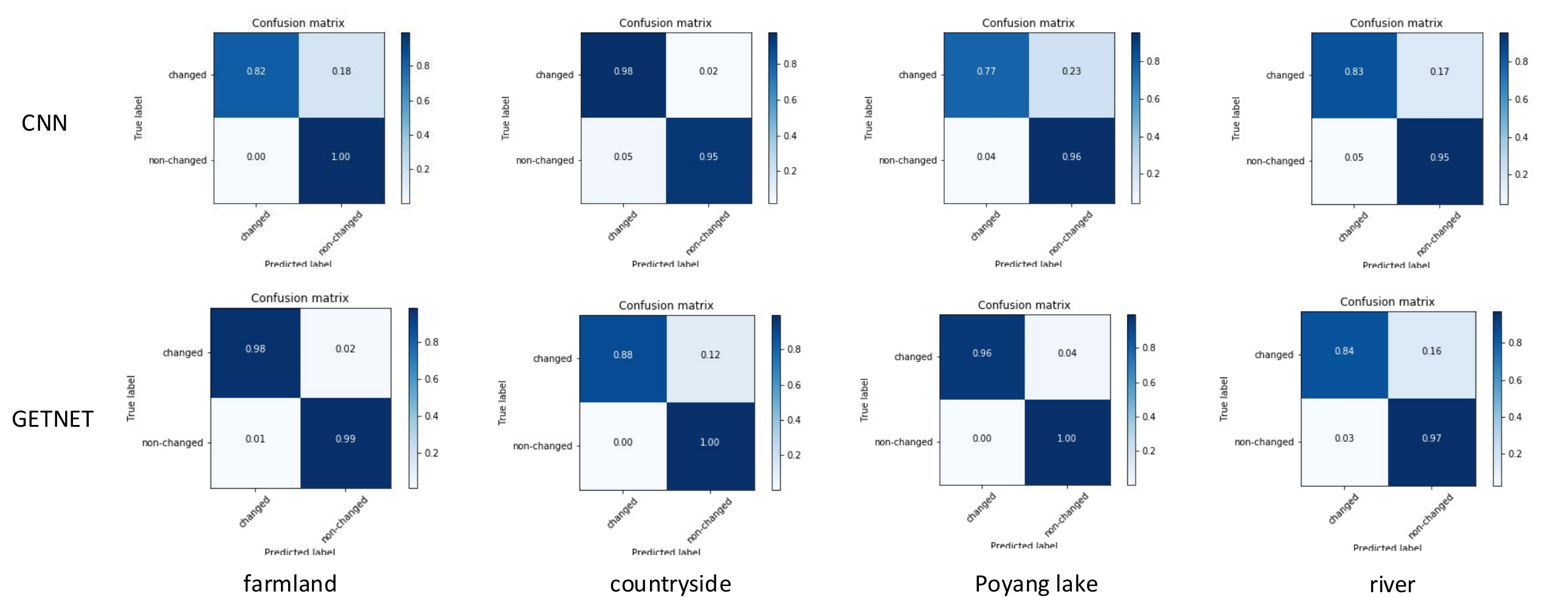}
	\caption{The confusion matrices of the patch-based CNN and GETNET.}
	\label{ConfuMatrix}
\end{figure*}

For the setup of experiments, the proposed GETNET is implemented by using Keras with Tensorflow as backend. The modified detail of the CNN network architecture is shown in Table \ref{tab1}. The GETNET is trained from scratch and no pre-trained models are used in this work. The batch size used for training is set to 96. The optimizer is Adagrad with epsilon ${10^{-8}}$ and initial learning rate is set to ${10^{-4}}$ without decay. The training parameter setups are the same on all data sets, which are trained with 30000 steps. The OA and Kappa coefficients of six CD methods on different data sets are shown in Table \ref{my-label}. To evaluate the proposed GETNET, we run 5 times per data set and get the mean value together with the standard deviation as the final CD results. The results in Table \ref{my-label} show that GETNET is robust across different data sets with the same parameter setting. The histogram of the OA comparisons on four data sets in different methods is shown in Fig. \ref{bar_OA}. The confusion matrices compare the performance of CNN and GETNET method in Fig. \ref{ConfuMatrix}. The comparisons of OA in GETNET and GETNET without unmixing are illustrated in Fig. \ref{ablation-fig}.

\subsubsection{\textbf{Experiments on the Farmland Data Set}} As shown in Fig. \ref{farmland}, the main change is the size of farmland on this data set. Table \ref{my-label} gives the details of different methods on four data sets. After overall consideration of OA and Kappa coefficient, it can be obviously found that GETNET is an effective method for CD task. With respect to OA, conventional CVA, PCA-CVA and IR-MAD perform better than CNN, but the performance of SVM is poor. Generally, SVM, CNN and GETNET use semi-supervised learning to train the classifier and network with the same training sets. Therefore, noise in pseudo training set can have negative effect on the accuracy of semi-supervised learning. SVM, as a binary classifier, divides the image into changed and unchanged areas but cannot deal adequately with noise. Moreover, CNN based deep learning is also not effective as expected, for it is more easily affected by noise in pseudo training set. As can be clearly seen, the best satisfactory performance is provided by the proposed GETNET, since it makes full use of hyperspectral data with abundance information and high anti-noise property. Even GETNET without unmixing works well and achieves the second-best performance. From another respect, the Kappa coefficient of GETNET is also high, which means the agreement between GETNET and the ground-truth map is almost perfect.  

\begin{figure}
	\centering
	\includegraphics[width=0.45\textwidth]{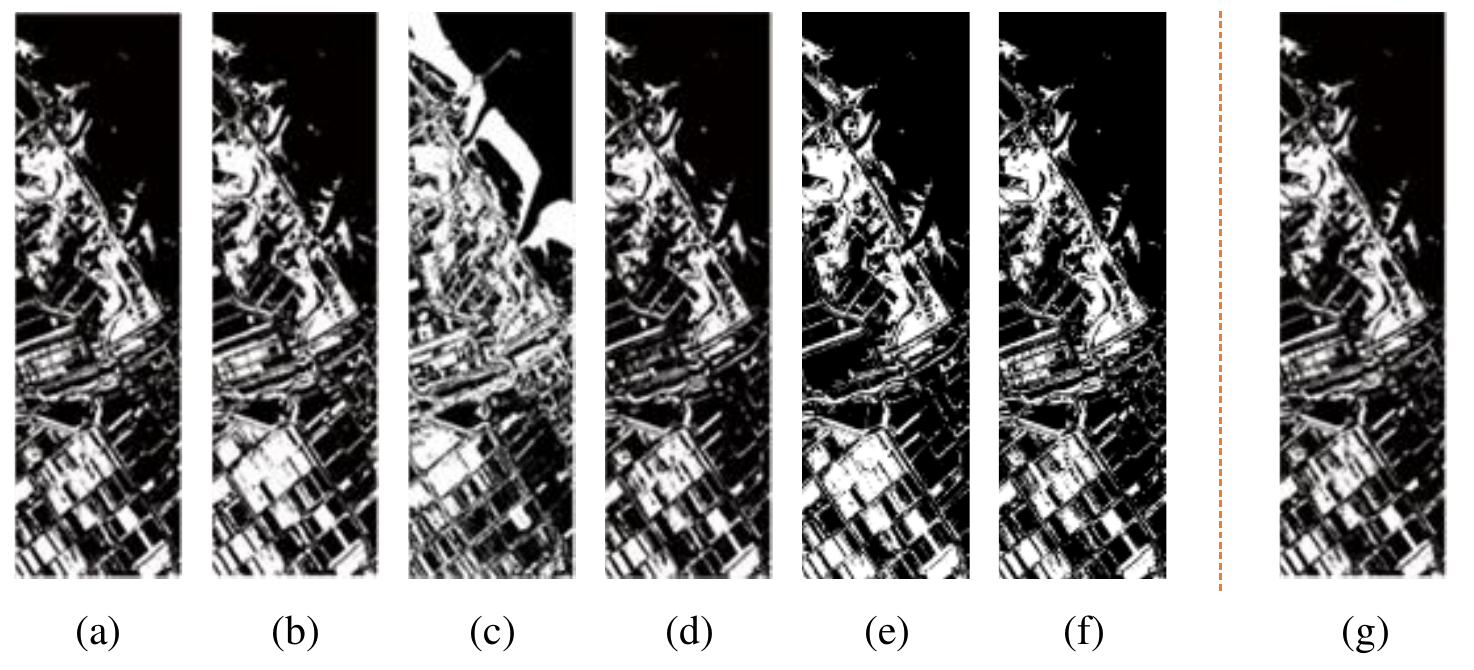}
	\caption{The change difference images of different methods on the data set of countryside, (a) the change difference image of CVA, (b) PCA-CVA, (c) IR-MAD, (d) SVM, (e) CNN, (f) GETNET, and (g) ground-truth map.}
	\label{countryside}
\end{figure}
\begin{figure}
	\centering
	\includegraphics[width=0.45\textwidth]{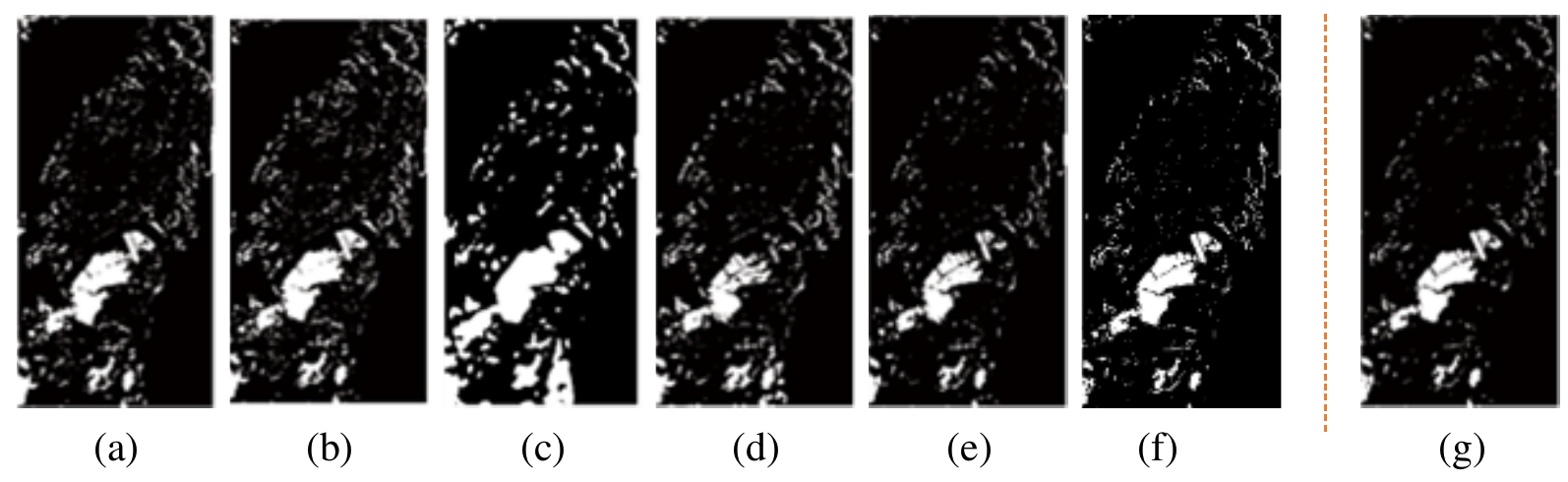}
	\caption{The change difference images of different methods on the data set of Poyang lake, (a) the change difference image of CVA, (b) PCA-CVA, (c) IR-MAD, (d) SVM, (e) CNN, (f) GETNET, and (g) ground-truth map.}
	\label{Poyang lake}
\end{figure}
\begin{figure}
	\centering
	\includegraphics[width=0.45\textwidth]{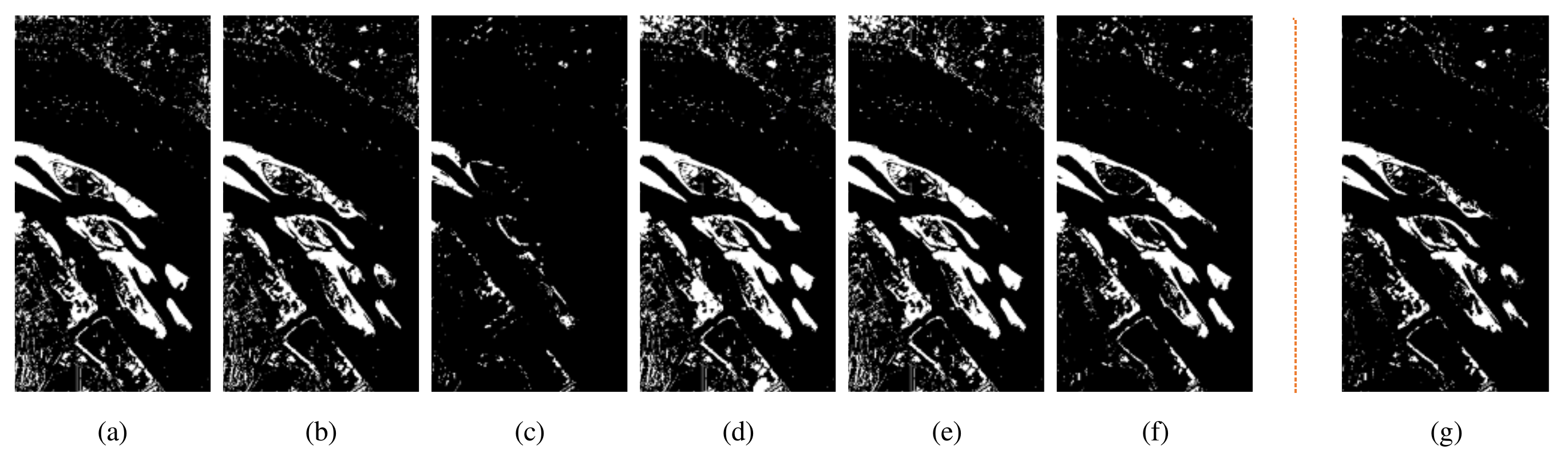}
	\caption{The change difference images of different methods on the data set of river, (a) the change difference image of CVA, (b) PCA-CVA, (c) IR-MAD, (d) SVM, (e) CNN, (f) GETNET, and (g) ground-truth map.}
	\label{river}
\end{figure}

\subsubsection{\textbf{Experiments on the Countryside Data Set}}The size of rural areas is the primary change and the changed information is quite complicated as shown in Fig. \ref{countryside}. In this data set, the best performance is GETNET. Specifically, IR-MAD yields lower OA, for it is not very sensitive to complicated changes. Because of dimension-reduction, PCA-CVA loses a part of information producing worse performance than CVA. Although the OA of CNN reaches more than 90\%, GETNET is appropriate 8\% higher. This is primarily because the inputs of them are different. The patch-based CNN only exploits the spatial context information of the preprocessed HSIs, while GETNET utilizes mixed-affinity matrix, which is an efficient way for simultaneous processing of hyperspectral data with abundance maps. It is worth noting that GETNET and CVA achieve a high OA of over 98\%. GETNET without unmixng yields lower OA than CVA, since pseudo training sets may be not sufficient to effectively train a deep network. Remarkably, abundance information provides subpixel level presentation, which improves the performance of CD. Though this data set is more complicated, both OA and Kappa coefficient of GETNET are even improved than other data sets.

\subsubsection{\textbf{Experiments on the Poyang Lake Data Set}} On this data set, there are some dispersed changes covering the lake. The performance of IR-MAD is not very satisfactory in terms of both OA and Kappa coefficient, for it cannot separate various changes well. CVA, PCA-CVA and SVM methods are relatively insensitive to changes in the width of the spacers. The Kappa coefficients of PCA-CVA, IR-MAD and SVM are also low. Though CNN and GETNET are both based on deep learning, their performances are different. As can be seen in Fig. \ref{Poyang lake}, some of obvious changes are not recognized well by CNN. However, GETNET is able to identify dispersed changes and changes in the width of the spacers. Overall, the best performance is generated by GETNET on this data set and GETNET without unmixing comes second at OA. 

\begin{figure*}
	\centering
	\includegraphics[width=1.0\textwidth]{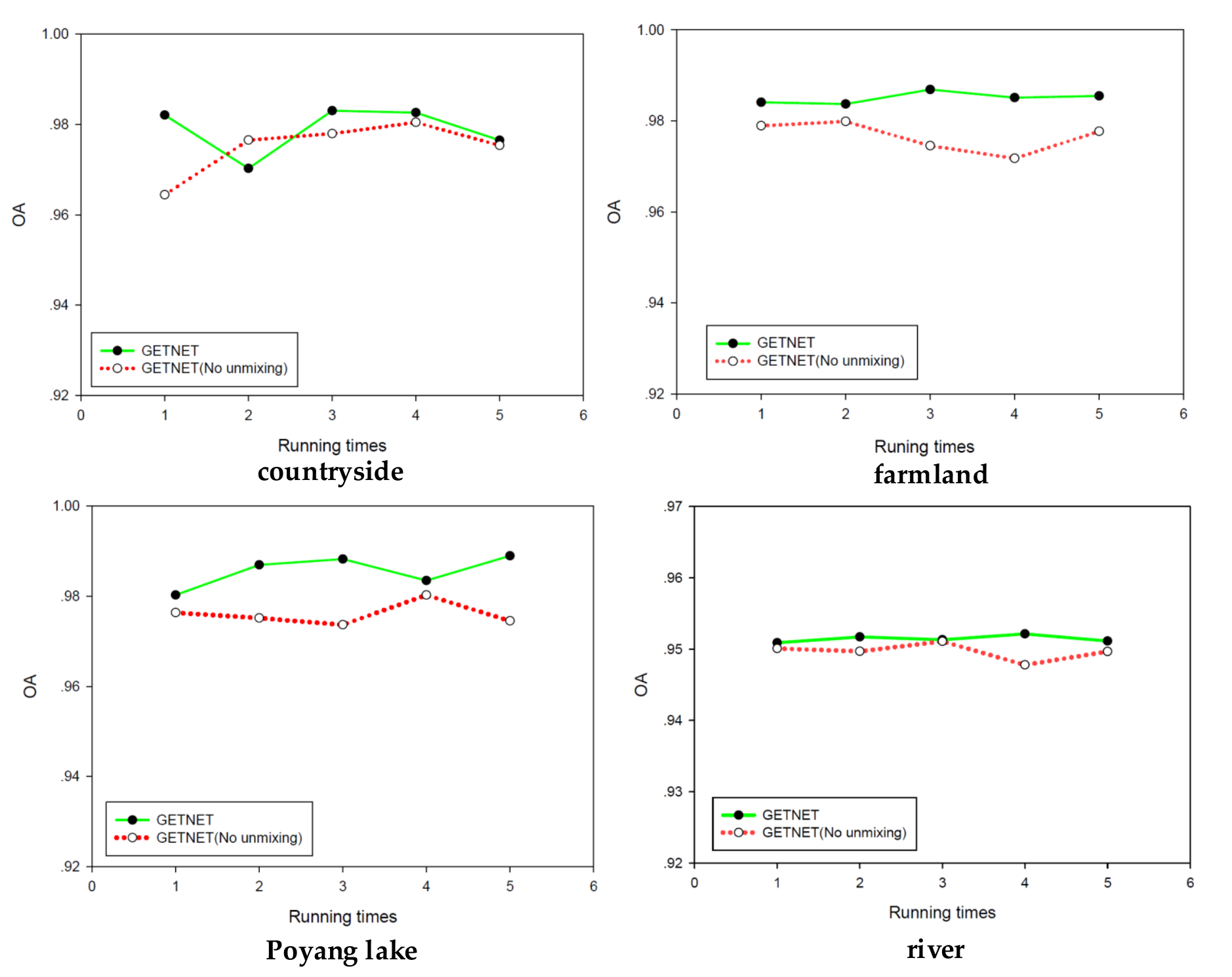}
	\caption{The comparisons of OA in GETNET and GETNET without unmixing}
	\label{ablation-fig}
\end{figure*}

\subsubsection{\textbf{Experiments on the River Data Set}}The disappearance of substance in river is an obvious change, and beyond that there are some dispersed changes. Fig. \ref{river} details the visual comparison of six different methods. In this dataset, the best result is CVA but not GETNET. It is thought that GETNET should always perform better than CVA because of the pseudo training sets generated by CVA. However, in a few data sets, these generated data sets may be insufficient to effectively train a deep network without parameter adjustment. The performances of CVA and GETNET are very close, both generating OA up to 95\%. PCA-CVA and CNN have the similar accuracy about 94\%. Besides, IR-MAD offers lower OA and Kappa coefficient, for it cannot identify irregular changes well. In addition, a remarkable problem is that the Kappa coefficient of each method is relatively lower than other data set. This is because the data set is more complicated and include diverse changes. It also demonstrates that the data set we create is reliable.

\subsubsection{\textbf{Ablation study of GETNET}} In this work, a mixed-affinity matrix is constructed as the input of 2-D CNN for discriminative features extraction. As illustrated in Fig. \ref{fig:aff}, abundance maps are used to provide extra subpixel information to improve the performance of CD. To validate the effectiveness of the unmixing information, we carry out the experiments of GETNET with and without abundance maps. The experimental results are shown in Fig. \ref{ablation-fig}, which demonstrate that GETNET is almost always better than GETNET without unmixing information across all the four data sets, since subpixel level representation is of good use to CD. Moreover, it can also be seen from the line chart that the performance of GETNET is stable and robust on different data sets and weight initials.

Although both GETNET and CNN are based on deep learning, the performance of GETNET is better according to Table \ref{my-label} and Fig. \ref{ConfuMatrix}. On the one hand, GETNET with end-to-end manner to train the deep neural network, can avoid error accumulation of some deep learning methods with pipeline procedure. On the other hand, the fusion of multi-sourced data, including hyperspectral data and unmixing abundance information, makes GETNET mine the characteristics of spectral gradient producing higher robustness, accuracy and generalization.

The proposed method outperforms the other CD methods in most of the cases and is robust for different HSIs. The data sets used in the experiments involve multiple change types, which demonstrate the merits of GETNET. The merits are mainly due to the novel mixed-affinity matrix with abundant information and unique design of GETNET.

\section{Conclusion}
\label{Conclusion}
In this paper, a general framework named GETNET is proposed for HSI-CD task by 2-D CNN. First, a novel mixed-affinity matrix is designed. It not only is an efficient way for simultaneous processing of multi-source information fusion, but also provides more abundant cross-channel gradient information. Next, based on deep learning, GETNET is developed to learn a series of more significant features fully mined, with excellent capabilities of generalization and robustness. Moreover, a new HSI-CD data set is designed for objective comparison of different methods. In the end, the proposed method is evaluated on existing and newly constructed data sets and shows its effectiveness through extensive comparisons and analyses.



\ifCLASSOPTIONcaptionsoff
  \newpage
\fi

\bibliographystyle{IEEEbib}
\bibliography{reference}

\begin{IEEEbiography}[{\includegraphics[width=1in,height=1.25in,clip,keepaspectratio]{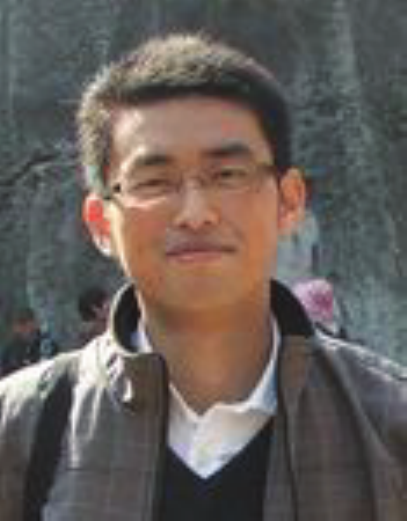}}]{Qi Wang} (M'15-SM'15) received the B.E. degree in automation and the Ph.D. degree in pattern recognition and intelligent systems from the University of Science and Technology of China, Hefei, China, in 2005  and 2010, respectively.  He is currently a Professor with the School of Computer Science, with the Unmanned System Research Institute, and with the Center for OPTical IMagery Analysis and Learning, Northwestern Polytechnical University, Xi'an, China. His research interests include computer vision and pattern recognition.
\end{IEEEbiography}
\begin{IEEEbiographynophoto}{Zhenghang Yuan}  received the B.E. degree in information security from the Northwestern Polytechnical University, Xi'an, China, in 2017. She is currently working toward the M.S. degree in computer science in the Center for OPTical IMagery Analysis and Learning, Northwestern Polytechnical University, Xi'an, China. Her research interests include hyperspectral image processing and computer vision.
\end{IEEEbiographynophoto}
\begin{IEEEbiographynophoto}{Qian Du} (S'98-M'00-SM'05-F'17) received the Ph.D. degree in electrical engineering from the University of Maryland–Baltimore County, Baltimore, MD, USA, in 2000. She is currently a Bobby Shackouls Professor with the Department of Electrical and Computer Engineering, Mississippi State University, Starkville, MS, USA. Her research interests include hyperspectral remote sensing image analysis and applications, pattern classification, data compression, and neural networks.
\end{IEEEbiographynophoto}
\begin{IEEEbiographynophoto}{Xuelong Li} (M'02-SM'07-F'12) is a Full Professor with the Center for Optical Imagery Analysis and Learning, Xi'an Institute of Optics and Precision
Mechanics, Chinese Academy of Sciences, Xi'an 710119, China.
\end{IEEEbiographynophoto}

\end{document}